\documentclass[twoside,11pt]{article}

\usepackage{graphicx}
\usepackage{caption}
\usepackage{apacite}
\begin{document}

\title{\textbf{Finding Original Image Of A Sub Image Using CNNs}}

\author{Raja Asim \\l154097@lhr.nu.edu.pk
	\\ FAST-NUCES Lahore}

\maketitle

\begin{abstract}
Convolututional Neural Networks have achieved state of the art in image classification, object detection and other image related tasks. In this paper I present another use of CNNs i.e. if given a set of images and then giving a single test image the network identifies that the test image is part of which image from the images given before. This is a task somehow similar to measuring image similarity and can be done using a simple CNN. Doing this task manually by looping can be quite a time consuming problem and won't be a generalizable solution. The task is quite similar to doing object detection but for that lots training data should be given or in the case of sliding window it takes lot of time and my algorithm can work with much fewer examples, is totally unsupervised and works much efficiently. Also, I explain that how unsupervised algorithm like K-Means or supervised algorithm like K-NN are not good enough to perform this task. The basic idea is that image encodings are collected for each image from a CNN, when a test image comes it is replaced by a part of original image, the encoding is generated using the same network, the frobenius norm is calculated and if it comes under a tolerance level then the test image is said to be the part of the original image.
\end{abstract}

\section{Related Work}
For image similarity \cite{radenovic2016cnn} the best known method so far is siamese network \cite{bromley1994signature} and for object detection and localization yolo \cite{redmon2016you} is mostly used. My work is somehow similar to both of these algorithms. KMeans \cite{hartigan1979algorithm} and KNN \cite{larose2005k} can also be used to achieve the same objective but in practice both of these algorithms' results did not go quite good.A lot of work has been done in object detection and finding similar images but no prior work has been done in finding the whole image whose part a test image might be.

\subsection{Siamese Network}
Siamese networks use CNNs \cite{krizhevsky2012imagenet} for one shot learning tasks \cite{fei2006one}. They have achieved near state of the art results in doing so. A siamese network consists of two networks that have an energy function on the top. The weights of the networks are shared and thus image encodings are generated for a picture and another picture is fed through the same network and when both the image encodings have been generated, the distance is computed and then the prediction is made that whether they belong to the same class or not. Notably no output unit is used and the encodings are collected from the fully connected layers at the end. Today siamese nets are being used in measuring image similarity extensively. \textit{My work is quite like this network but the distinction of my network from the siamese network is that instead of predicting classes the image encodings of the image is matched with the same image but by switching test image with a piece of the original image to see if it is part of original image or not. Also my network is unsupervised.} A traditional siamese network can't be used here because the encodings of a similar image might be close to the original image rather than the part of the image i.e. test image.

\subsection{Yolo}
Yolo algorithm is used for object detection and localization. It makes bounding boxes around the objects to detect and also outputs the class probabilities. It is so efficient and can detect objects at real time. It is made up of a deep CNN. The initial convolututional and pooling layers detect features and the fully connected layers at the end predicts bounding boxes and the probabilities. \textit{The main difference between my network and Yolo is that Yolo learns to predict bounding boxes from the training data but since my algorithm is unsupervised so no training data is needed. Also Yolo is used for object detection which is somehow similar to the "finding an image whose part the test image is" task but the sub images i.e. test images can be a little bit different and Yolo doesn't generalize well on a little deviated examples. }  

\subsection{Problem with Kmeans and KNN}
The main problem that occurred in practice using KMeans for this task was that if there are very much similar images in the dataset then they were added to the same cluster. The value of K was set to number of images present in the dataset. So in the ideal case the test image had to fall into the same cluster with one of the dataset image i.e. of whom it is a part. But what happened was that if some images are same in the dataset then they were added into the same cluster and test image was added to a separate cluster alone. On any other value of K for the Kmeans the results were horrible because test image was being added in different clusters on each iteration of Kmeans and similar images i.e. in color from the dataset were added to the same cluster.
\par
The nearest neighbors set for KNN were 2 i.e. one actual image and one any other most near image. The most near image obtained is on the basis of a distance measure like L2-norm or L1-norm. But this does not tell anything about the image being part or not. It depends on the RGB values of the image, no shapes, no edges or image features are taken under consideration. Instead a CNN detects features from the convolututional layer and keeps the most important one in the max pooling. Intuitively using CNN is a better approach and in practice it did quite well. KMeans and KNN both did not do a good job for this particular problem.

\section{Introduction}
There are lots of algorithms available for supervised object detection but till now no such algorithm will the answer the question like if there are some unlabeled images and one has to tell whether a test image is a part of any of these images and if so then which one is it. The primary motivation for this work is that a lot of work has been done in computer vision tasks but no such algorithm or method exists which can get this job done. Siamese network is the closest to address the same problem but if two pictures are similar we can't guarantee that one is the part of other it might be or might be not. A simple CNN that finds original image of a test image can be used a lot in several cases. It is quite a generalizable solution as one needs to change the tolerance level and the desired results will be produced.

\section{Network Architecture}
The network contained three convolututional layers each of them followed by a max pooling layer \cite{masci2011stacked}. The convolututional layers have filters in the format of multiple of 2 and starting from 16 i.e. 16,32,64,128,256 filter size was 5x5 and the activation function used was leaky relu \cite{2015arXiv150500853X}. All the weights were initialized with the standard deviation of 0.01 and mean of 0.001 in first convolututional layer, 0.02 and 0.002 in the second layer, 0.03 and 0.003 in the third layer and so on. The max pooling layer was used for down sampling and had the filter size of 5x5 and the strides were used in the order of 2,1,2,1 and so on. The output layer just had 10 neurons and the activation here also was leaky relu. Notably no fully connected layers were used and all the weights of each layer were different in mean and standard deviation than the other layer's weights. This actually helped in capturing better encodings because since each image had different pixel values and unique weights are being used, so the encodings of the unique images were always quite diverse and different but encodings of similar images were somewhat close.

\subsection{Methodology}
For all the images present in the dataset the encodings were generated with just one forward pass. So, no weight updating happened and no optimizer was used. Because the main objective is to just relate the image pixels to some meaningful encodings. After all the initial images' encodings have been generated, a test image was taken. Resized to 64x64x3 and 4 pieces were made of the first image from the dataset. The equal pieces made were of 64x64x3 size. So a 128x128x3 image was broken down into upper right, upper left, lower right and lower left images. Just one piece was replaced with the test image at one iteration and the image was recombined in such a manner that always 3 image pieces are of the original image and 1 is of test image. So one can see that to perform task of finding whole image of a sub-image this is the most intuitive solution. This new image obtained is then passed through the above explained same network and some encodings were generated. The test image is replaced by all pieces one by one and this process continues till the end of all images present in the dataset and the encodings are thus collected. The process was a little bit slow because images were being sliced and then test image replaced a slice.

\subsection{Difference between encodings}
Frobenius norm \cite{bay2006surf} was used as a distance measure between the generated image and the original image. The tolerance level was set to 100 i.e. if the difference was more than 100 the test image is not part of the original image from the dataset. The tolerance could have been set lower but since the test image can be oriented in many ways other than the original piece so it caused some deviation between encodings, so to adjust that the tolerance level was set somehow high. But whenever a different image came since all the weights were different and the difference rises up which gives a strong evidence that this image can't be part of the other image. Tolerance level can be set lower if the test image is exactly a quarter equal to the original image but since in practice a test image can be of greater or lower size so that's why tolerance level needs to be higher. Also a higher tolerance level means that if test image is somehow similar to the original image but not a part of it then it will also be selected.
\par
The test image was merged with the original image in 4 different ways i.e. in upper right, upper left, lower right and lower left position. So at any time there were 3 original image parts and 1 part was of generated image. After getting the encodings, the frobenius norm was calculated between original image's encodings and the 4 generated images' encodings. If any one of the norm was below or equal to the tolerance level, the test image was called as the part of the original image.

\section{Results and Conculsion}
I mainly tested this network on screenshots as the dataset was easily made from a simple nodejs application that takes screenshots after several predefined seconds and on the Flickr 8k images dataset \cite{flickr}. The encodings of all the dataset images were generated and saved, then the test images' encodings were generated by replcaing them with a piece of original image from dataset and the frobenius norm was calculated to measure the difference between encodings. The process of testing was a little bit slow on a core i3 laptop i.e. for a dataset of 4000 images and for one test image it took nearly 5 minutes to tell that which picture's part the test image is. Some of the pictures of the dataset, test pictures and the generated pictures for encoding matching are given below:-

\begin{figure}
	\centering
	\begin{minipage}{.5\textwidth}
		\centering
		\includegraphics[width=40mm]{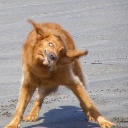}
		\captionof{figure}{Dataset image}
		\label{fig:test1}
	\end{minipage}%
	\begin{minipage}{.5\textwidth}
		\centering
		\includegraphics[width=40mm]{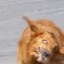}
		\captionof{figure}{Test image}
		\label{fig:test2}
	\end{minipage}
\end{figure}

\begin{figure}
	\centering
	\begin{minipage}{.5\textwidth}
		\centering
		\includegraphics[width=40mm]{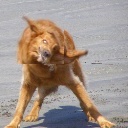}
		\captionof{figure}{Generated Image 1}
		\label{blah}
	\end{minipage}%
	\begin{minipage}{.5\textwidth}
		\centering
		\includegraphics[width=40mm]{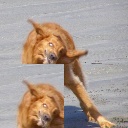}
		\captionof{figure}{Generated Image 2}
		\label{fig:test4}
	\end{minipage}
	\begin{minipage}{.5\textwidth}
		\centering
		\includegraphics[width=40mm]{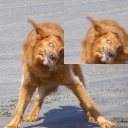}
		\captionof{figure}{Generated Image 3}
		\label{fig:test5}
	\end{minipage}%
	\begin{minipage}{.5\textwidth}
		\centering
		\includegraphics[width=40mm]{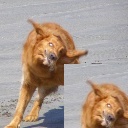}
		\captionof{figure}{Generated Image 4}
		\label{fig:test6}
	\end{minipage}
\end{figure}

\subsection{Testing on flickr dataset}
The dataset contained 4000 different images i.e. images were of no particular category, there were dogs, cats, trees, cars, rocks etc all different images were present. All images were resized to 128x128x3. A single image was selected, cropped from a random point and was resized to 64x64x3. Now according to the above described procedure these were the results I got:-

\begin{table}[ht]
	\caption{Results of Classification}
	
	\centering 
	\begin{tabular}{c c c c c}
		\hline \hline                        %inserts double horizontal lines
		Dataset & Test Images & Accuracy \\ [0.5ex]
		\hline
		4000& 100 &  95.5\%\\
		
		\hline
	\end{tabular}
	\label{table:nonlin}
\end{table}

The accuracy was calculated by dividing test images predicted correctly as part of their original image and total testing iterations done i.e. 100 in my case and multiplying the answer by 100 to get a percentage.

\section{Working on low image sizes}
If the test image is low sized the intuitive solution is to slice the original image also in smaller pieces and then replace a piece with the test image. This worked perfectly in practice and then 16 pieces of 32x32x3 were made of the original image and 1 test piece of 32x32x3 was replaced and after generating encodings, the frobenius norm was calculated and the tolerance was set a bit higher i.e. to 200 because the larger the slices, the different it becomes from the original image with a small change. So to adjust flexibility to the generated image, the tolerance level was set higher. Also, this process took more time as the slices were more and instead of 4 image encodings now there were 16 per image.

\section{Drawbacks and Limitations}
One clear limitation of my method is that we need to have a prior knowledge of the test image size. If the test image is just a tiny portion of original image then making just 4 slices does not make any sense as the encodings will not be generated properly and the difference will be huge. Another drawback of the method is that it takes a lot of time. It takes less time than a human but still 5 minutes per picture are huge. Maybe a more sophisticated algorithm can be devised for this problem which takes a lot less time. 

\section{Future Work}
The future work includes to see how the same task can be achieved in a less amount of time. 5 minutes for one testing picture is a lot of time. Also as the test image size decreases the process becomes a lot lengthy i.e. generating 16 images instead of 4 for each image etc. So devising a sophisticated algorithm for this task is included in my future work aspirations. Also currently if a test image is similar to the original image it is chosen because the tolerance level is high so my future work also includes addressing this problem that only part of image is picked up and not any other similar image.
 
\bibliographystyle{apacite}
\bibliography{references}

\begin{thebibliography}{}

\bibitem [\protect \citeauthoryear {%
Bay%
, Tuytelaars%
\BCBL {}\ \BBA {} Van~Gool%
}{%
Bay%
\ \protect \BOthers {.}}{%
{\protect \APACyear {2006}}%
}]{%
bay2006surf}
\APACinsertmetastar {%
bay2006surf}%
\begin{APACrefauthors}%
Bay, H.%
, Tuytelaars, T.%
\BCBL {}\ \BBA {} Van~Gool, L.%
\end{APACrefauthors}%
\unskip\
\newblock
\APACrefYearMonthDay{2006}{}{}.
\newblock
{\BBOQ}\APACrefatitle {Surf: Speeded up robust features} {Surf: Speeded up
  robust features}.{\BBCQ}
\newblock
\BIn{} \APACrefbtitle {European conference on computer vision} {European
  conference on computer vision}\ (\BPGS\ 404--417).
\PrintBackRefs{\CurrentBib}

\bibitem [\protect \citeauthoryear {%
Bromley%
, Guyon%
, LeCun%
, S{\"a}ckinger%
\BCBL {}\ \BBA {} Shah%
}{%
Bromley%
\ \protect \BOthers {.}}{%
{\protect \APACyear {1994}}%
}]{%
bromley1994signature}
\APACinsertmetastar {%
bromley1994signature}%
\begin{APACrefauthors}%
Bromley, J.%
, Guyon, I.%
, LeCun, Y.%
, S{\"a}ckinger, E.%
\BCBL {}\ \BBA {} Shah, R.%
\end{APACrefauthors}%
\unskip\
\newblock
\APACrefYearMonthDay{1994}{}{}.
\newblock
{\BBOQ}\APACrefatitle {Signature verification using a" siamese" time delay
  neural network} {Signature verification using a" siamese" time delay neural
  network}.{\BBCQ}
\newblock
\BIn{} \APACrefbtitle {Advances in Neural Information Processing Systems}
  {Advances in neural information processing systems}\ (\BPGS\ 737--744).
\PrintBackRefs{\CurrentBib}

\bibitem [\protect \citeauthoryear {%
Fei-Fei%
, Fergus%
\BCBL {}\ \BBA {} Perona%
}{%
Fei-Fei%
\ \protect \BOthers {.}}{%
{\protect \APACyear {2006}}%
}]{%
fei2006one}
\APACinsertmetastar {%
fei2006one}%
\begin{APACrefauthors}%
Fei-Fei, L.%
, Fergus, R.%
\BCBL {}\ \BBA {} Perona, P.%
\end{APACrefauthors}%
\unskip\
\newblock
\APACrefYearMonthDay{2006}{}{}.
\newblock
{\BBOQ}\APACrefatitle {One-shot learning of object categories} {One-shot
  learning of object categories}.{\BBCQ}
\newblock
\APACjournalVolNumPages{IEEE transactions on pattern analysis and machine
  intelligence}{28}{4}{594--611}.
\PrintBackRefs{\CurrentBib}

\bibitem [\protect \citeauthoryear {%
Hartigan%
\ \BBA {} Wong%
}{%
Hartigan%
\ \BBA {} Wong%
}{%
{\protect \APACyear {1979}}%
}]{%
hartigan1979algorithm}
\APACinsertmetastar {%
hartigan1979algorithm}%
\begin{APACrefauthors}%
Hartigan, J\BPBI A.%
\BCBT {}\ \BBA {} Wong, M\BPBI A.%
\end{APACrefauthors}%
\unskip\
\newblock
\APACrefYearMonthDay{1979}{}{}.
\newblock
{\BBOQ}\APACrefatitle {Algorithm AS 136: A k-means clustering algorithm}
  {Algorithm as 136: A k-means clustering algorithm}.{\BBCQ}
\newblock
\APACjournalVolNumPages{Journal of the Royal Statistical Society. Series C
  (Applied Statistics)}{28}{1}{100--108}.
\PrintBackRefs{\CurrentBib}

\bibitem [\protect \citeauthoryear {%
Krizhevsky%
, Sutskever%
\BCBL {}\ \BBA {} Hinton%
}{%
Krizhevsky%
\ \protect \BOthers {.}}{%
{\protect \APACyear {2012}}%
}]{%
krizhevsky2012imagenet}
\APACinsertmetastar {%
krizhevsky2012imagenet}%
\begin{APACrefauthors}%
Krizhevsky, A.%
, Sutskever, I.%
\BCBL {}\ \BBA {} Hinton, G\BPBI E.%
\end{APACrefauthors}%
\unskip\
\newblock
\APACrefYearMonthDay{2012}{}{}.
\newblock
{\BBOQ}\APACrefatitle {Imagenet classification with deep convolutional neural
  networks} {Imagenet classification with deep convolutional neural
  networks}.{\BBCQ}
\newblock
\BIn{} \APACrefbtitle {Advances in neural information processing systems}
  {Advances in neural information processing systems}\ (\BPGS\ 1097--1105).
\PrintBackRefs{\CurrentBib}

\bibitem [\protect \citeauthoryear {%
Larose%
}{%
Larose%
}{%
{\protect \APACyear {2005}}%
}]{%
larose2005k}
\APACinsertmetastar {%
larose2005k}%
\begin{APACrefauthors}%
Larose, D\BPBI T.%
\end{APACrefauthors}%
\unskip\
\newblock
\APACrefYearMonthDay{2005}{}{}.
\newblock
{\BBOQ}\APACrefatitle {k-nearest neighbor algorithm} {k-nearest neighbor
  algorithm}.{\BBCQ}
\newblock
\APACjournalVolNumPages{Discovering knowledge in data: An introduction to data
  mining}{}{}{90--106}.
\PrintBackRefs{\CurrentBib}

\bibitem [\protect \citeauthoryear {%
Masci%
, Meier%
, Cire{\c{s}}an%
\BCBL {}\ \BBA {} Schmidhuber%
}{%
Masci%
\ \protect \BOthers {.}}{%
{\protect \APACyear {2011}}%
}]{%
masci2011stacked}
\APACinsertmetastar {%
masci2011stacked}%
\begin{APACrefauthors}%
Masci, J.%
, Meier, U.%
, Cire{\c{s}}an, D.%
\BCBL {}\ \BBA {} Schmidhuber, J.%
\end{APACrefauthors}%
\unskip\
\newblock
\APACrefYearMonthDay{2011}{}{}.
\newblock
{\BBOQ}\APACrefatitle {Stacked convolutional auto-encoders for hierarchical
  feature extraction} {Stacked convolutional auto-encoders for hierarchical
  feature extraction}.{\BBCQ}
\newblock
\BIn{} \APACrefbtitle {International Conference on Artificial Neural Networks}
  {International conference on artificial neural networks}\ (\BPGS\ 52--59).
\PrintBackRefs{\CurrentBib}

\bibitem [\protect \citeauthoryear {%
M.~Hodosh%
}{%
M.~Hodosh%
}{%
{\protect \APACyear {2013}}%
}]{%
flickr}
\APACinsertmetastar {%
flickr}%
\begin{APACrefauthors}%
M.~Hodosh, J\BPBI H., P.~Young.%
\end{APACrefauthors}%
\unskip\
\newblock
\APACrefYearMonthDay{2013}{}{}.
\newblock
{\BBOQ}\APACrefatitle {Framing Image Description as a Ranking Task: Data,
  Models and Evaluation Metrics} {Framing image description as a ranking task:
  Data, models and evaluation metrics}.{\BBCQ}
\newblock
\BIn{} (\BVOL~47).
\PrintBackRefs{\CurrentBib}

\bibitem [\protect \citeauthoryear {%
Radenovi{\'c}%
, Tolias%
\BCBL {}\ \BBA {} Chum%
}{%
Radenovi{\'c}%
\ \protect \BOthers {.}}{%
{\protect \APACyear {2016}}%
}]{%
radenovic2016cnn}
\APACinsertmetastar {%
radenovic2016cnn}%
\begin{APACrefauthors}%
Radenovi{\'c}, F.%
, Tolias, G.%
\BCBL {}\ \BBA {} Chum, O.%
\end{APACrefauthors}%
\unskip\
\newblock
\APACrefYearMonthDay{2016}{}{}.
\newblock
{\BBOQ}\APACrefatitle {CNN image retrieval learns from BoW: Unsupervised
  fine-tuning with hard examples} {Cnn image retrieval learns from bow:
  Unsupervised fine-tuning with hard examples}.{\BBCQ}
\newblock
\BIn{} \APACrefbtitle {European Conference on Computer Vision} {European
  conference on computer vision}\ (\BPGS\ 3--20).
\PrintBackRefs{\CurrentBib}

\bibitem [\protect \citeauthoryear {%
Redmon%
, Divvala%
, Girshick%
\BCBL {}\ \BBA {} Farhadi%
}{%
Redmon%
\ \protect \BOthers {.}}{%
{\protect \APACyear {2016}}%
}]{%
redmon2016you}
\APACinsertmetastar {%
redmon2016you}%
\begin{APACrefauthors}%
Redmon, J.%
, Divvala, S.%
, Girshick, R.%
\BCBL {}\ \BBA {} Farhadi, A.%
\end{APACrefauthors}%
\unskip\
\newblock
\APACrefYearMonthDay{2016}{}{}.
\newblock
{\BBOQ}\APACrefatitle {You only look once: Unified, real-time object detection}
  {You only look once: Unified, real-time object detection}.{\BBCQ}
\newblock
\BIn{} \APACrefbtitle {Proceedings of the IEEE conference on computer vision
  and pattern recognition} {Proceedings of the ieee conference on computer
  vision and pattern recognition}\ (\BPGS\ 779--788).
\PrintBackRefs{\CurrentBib}

\bibitem [\protect \citeauthoryear {%
{Xu}%
, {Wang}%
, {Chen}%
\BCBL {}\ \BBA {} {Li}%
}{%
{Xu}%
\ \protect \BOthers {.}}{%
{\protect \APACyear {2015}}%
}]{%
2015arXiv150500853X}
\APACinsertmetastar {%
2015arXiv150500853X}%
\begin{APACrefauthors}%
{Xu}, B.%
, {Wang}, N.%
, {Chen}, T.%
\BCBL {}\ \BBA {} {Li}, M.%
\end{APACrefauthors}%
\unskip\
\newblock
\APACrefYearMonthDay{2015}{{\APACmonth{05}}}{}.
\newblock
{\BBOQ}\APACrefatitle {{Empirical Evaluation of Rectified Activations in
  Convolutional Network}} {{Empirical Evaluation of Rectified Activations in
  Convolutional Network}}.{\BBCQ}
\newblock
\APACjournalVolNumPages{ArXiv e-prints}{}{}{}.
\PrintBackRefs{\CurrentBib}

\end{thebibliography}

\end{document}